# ANALYSIS OF OPINIONATED TEXT FOR OPINION MINING

K Paramesha[1] and K C Ravishankar[2]

[1]Department of Computer Engineering, Vidyavardhaka College of Engg., Mysuru, India
paramesha.k@vvce.ac.in
[2]Department of Computer Engineering, Government Engineering College, Hassan, India
kcr@gechassan.ac.in

## ABSTRACT

*In sentiment analysis, the polarities of the opinions expressed on an object/feature are determined to assess the sentiment of a sentence or document whether it is positive/negative/neutral. Naturally, the object/feature is a noun representation which refers to a product or a component of a product, let's say, the "lens" in a camera and opinions emanating on it are captured in adjectives, verbs, adverbs and noun words themselves. Apart from such words, other meta-information and diverse effective features are also going to play an important role in influencing the sentiment polarity and contribute significantly to the performance of the system. In this paper, some of the associated information/meta-data are explored and investigated in the sentiment text. Based on the analysis results presented here, there is scope for further assessment and utilization of the meta-information as features in text categorization, ranking text document, identification of spam documents and polarity classification problems.*

## 1. INTRODUCTION

Sentiment Analysis (SA) is a multifaceted problem [1, 2], which needs thorough understanding of the sentiment text syntactically and semantically [3] to be able to judge the sentiment polarities. Being a classification task, the machine learning approaches map sentiment text into feature vector which requires the extraction of several types of discriminating features. The meta-information associated with the text are also explored and exploited as features in many machine learning approaches [4, 5, 6, 7, 8]. In a paper presented by [9], highlights the explicit and implicit features beyond words themselves. For example, word's part-of-speech (POS) and its position considered to be the explicit one. The similarity score of a word with the polarity class seed word "good" is an implicit one. In the papers presented by [10, 11, 12], different types of features such as word features, structure features, sentence features etc. were explored. In fact there has been a broad range of features and methods [13, 14] that many researchers have been investigating and exploiting new ones for the sentiment classification and to enhance the performance of the system.

## 2. DATASET

The dataset [15] on which we analysed, consists of 294 reviews sampled across the domains books, dvds, music, electronics and videogames and has document sentiment categories positive, neutral and negative. The document reviews are annotated with the categories POS, NEG, NEU, MIX, and NR. The MIX and NR are combined with NEU and reduced to three classes. The document and sentence distribution are as shown in the Table 1.

Table 1. Review documents and sentences distribution.

| Category | Documents | | | | Sentences | | | |
|---|---|---|---|---|---|---|---|---|
| | **POS** | **NEG** | **NEU** | **Total** | **POS** | **NEG** | **NEU** | **Total** |
| Books | 19 | 20 | 20 | 59 | 160 | 195 | 384 | 739 |
| Dvds | 19 | 20 | 20 | 59 | 164 | 264 | 371 | 799 |
| Electronics | 19 | 19 | 19 | 57 | 161 | 240 | 227 | 628 |
| Music | 20 | 20 | 19 | 59 | 183 | 179 | 276 | 638 |
| Videogames | 20 | 20 | 20 | 60 | 255 | 442 | 335 | 1,032 |
| Total | 97 | 99 | 98 | 294 | 923 | 1,320 | 1,593 | 3,836 |

## 3. Experiment Setup

For computing the WordNet domains of words in the review sentences in which each word has many senses across POS in the WordNet, it is appropriate to find the right senses of words based on which WordNet domains are assigned rather than arbitrarily assign the WordNet domains based on the POS. The word sense disambiguation (WSD) is performed for each sentence using the services offered by the **idilia**[1] *(http://www.idilia.com)*. To get our text sense tagged, java application is implemented to communicate through REST services. To derive the vectors representing the six types of emotions for each document, similarity scores of the words with all the emotions types are computed with the aid of WordNet based word-similarity algorithms.

## 4. Features

To aid the SA using machine learning approaches, several associated linguistic and statistical features at sentence level and document level could be integrated [16, 17]. Many researches have been conducted to evaluate the performance of the features but a holistic feature set for efficient and high performance SA is hard to find. The aim of the work is to explore features and techniques in the input text data, which can contribute to the performance of the system.

### 4.1. Inter-Sentence Coherence

In a review document consisting of several sentences, the polarity of the document could be determined as the polarity with maximum votes of all the sentences. Intuitively, the probability of a sentence polarity in the review document is same as the document's polarity. But, it is also in influenced by the previous sentence's polarity. The probability distribution given in Table 2 shows the polarity probabilities of next sentence given the polarity of the current sentence across all the review documents and domains. It is interesting to know that the likelihood of the polarity of a sentence is the polarity of its previous sentence (of course not applicable to the first sentence of the review document). It is also noticeable that the probability of polarity transition to its opposite polarity (pos to **neg** and **neg** to **pos**) is lesser than the transition to neutral (**pos** to **neu** and **neg** to **neu**). With this understanding of the probability distribution, it implies that the abrupt transition in polarities of sentences is less likely to happen and while modelling the feature vector for the current sentence, the incorporation of features of its previous sentence is indeed going to contribute significantly to the performance of the system.

Table 2. Probability distribution of inter sentence polarities.

| | Next Sentence | | | |
|---|---|---|---|---|
| | **Polarity** | **Pos** | **Neg** | **Neu** |
| **Current Sentence** | **Pos** | **0.148** | 0.027 | 0.063 |
| | **Neg** | 0.018 | **0.234** | 0.095 |
| | **Neu** | 0.073 | 0.094 | **0.248** |

## 4.2. Emotions

It is known that SA is hard unless the whole sense of the words and phrases are known properly. A sentence can be positive or negative even though it doesn't contain any sentiment words and vice versa is also true. It is observed that six types of emotions {**Anger, Disgust, Fear, Joy, Sadness** and **Surprise**} can be mapped into polarities of the sentiments. As investigated in the paper [18], the emotions are tagged to the words in short sentences and then valence (**positive** and **negative**) is determined. For tagging, they have used enriched labels from the seed WordNet-Affect labels. But, we took rather a different approach in computing the emotions of the text. We employed the WordNet based word-similarity algorithms to compute and aggregate the similarity score of words with all of the emotional words. We analysed emotions both at sentence and document level. The emotion distributions over **pos**, **neg** and **neu** documents are plotted in the Figure 1 which shows some variation in each of the distribution between **pos** and **neg**. The distribution of emotion over **neu** documents is seen to be occupying the space between **pos** and **neg** distribution. Further, we tested the combination of all the emotion distributions on various binary classifiers which yielded an accuracy of 60% at the sentence level and 75% at document level.

## 4.3. WordNet Syntactic Domains

Many researchers have been actively working on using WordNet domains in computational linguistics. The WordNet domains contains around 200 domains labels such as economics, politics, law, science etc., which are organized in a hierarchical structure (WordNet Domains Hierarchy). Each synset of WordNet was labelled with one or more labels which could be explored to establish semantic relations among word senses and can be effectively utilized during disambiguation process. The domain information utility is realized in the applications such as Domain Driven Disambiguation (DDD) [19], word sense disambiguation (WSD) and text categorization (TC) [20]. Akin to the WordNet domains, there is another category of information associated with the WordNet synsets, which are organized into forty five lexicographer les based on syntactic category and logical grouping as shown in the Table 3. We have plotted the syntactic domains distribution over POS, NEG and NEU documents across fiv domains of the dataset after WSD process to investigate any potential use in deciding the polarity of sentences. The distributions of syntactic domains are shown in Figure 2. We could figure out that in electronics domain the **noun.artifact** is predominant compare to other domains whereas **noun.act**, **noun.cognition**, **noun.communication**, **noun.person**, **verb.cognition** and **verb.communication** have relatively significant distribution in all domains. The main idea for looking into the syntactic domains is that the domain information make-up a part of the information required for the word sense computations. Since word polarity depends on the word sense, it could be perceived that domain information have in influence on the polarity of words. Consider Sentence S1, which is POS polarity but doesn't contain any sentiment words. In this case, we can able to predict the polarity if we can establish the word "**buy**" belonging to **verb.possession** domain is inclined to positive sentiment.

**S1:** Netgear software makes me **buy** them over and over...

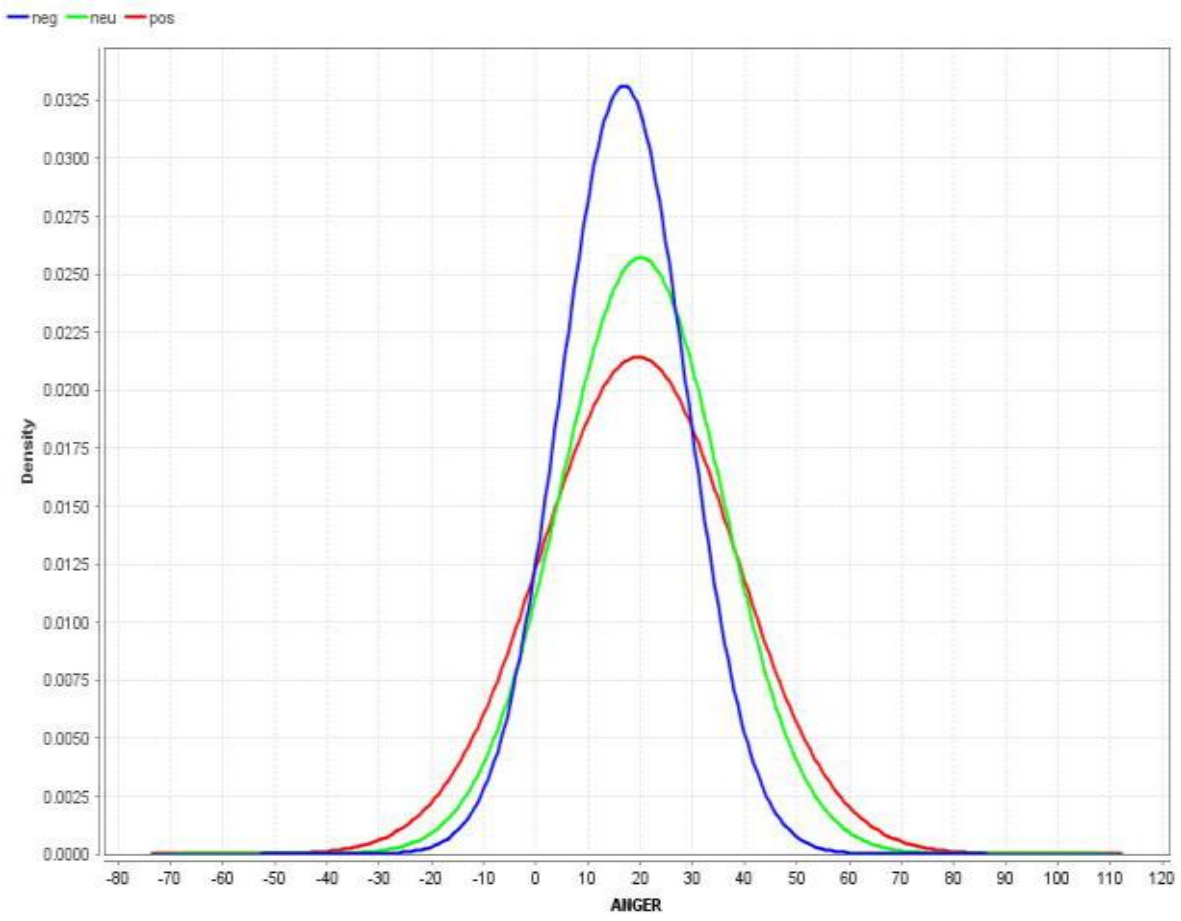

(a) ANGER

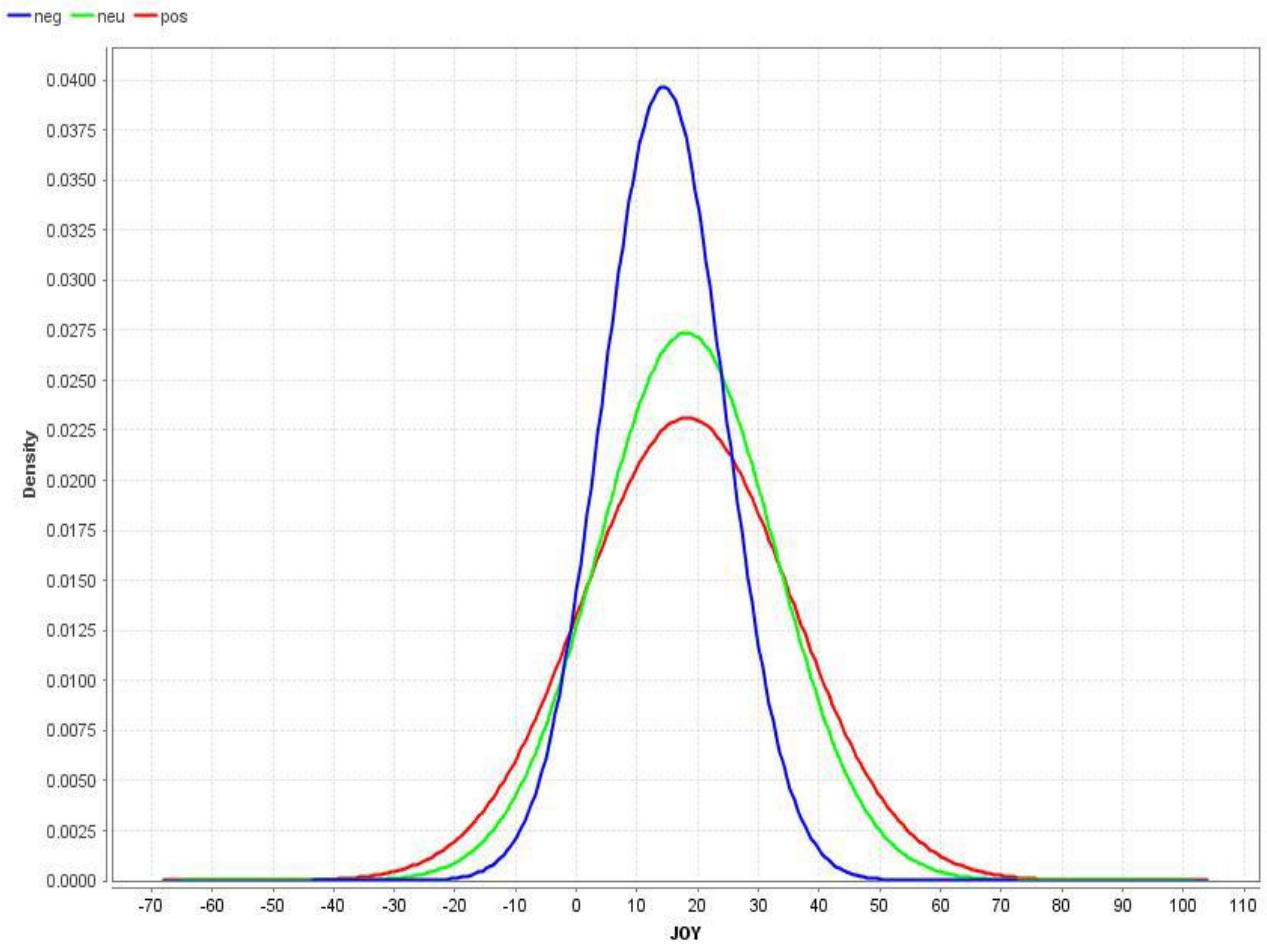

(b) JOY

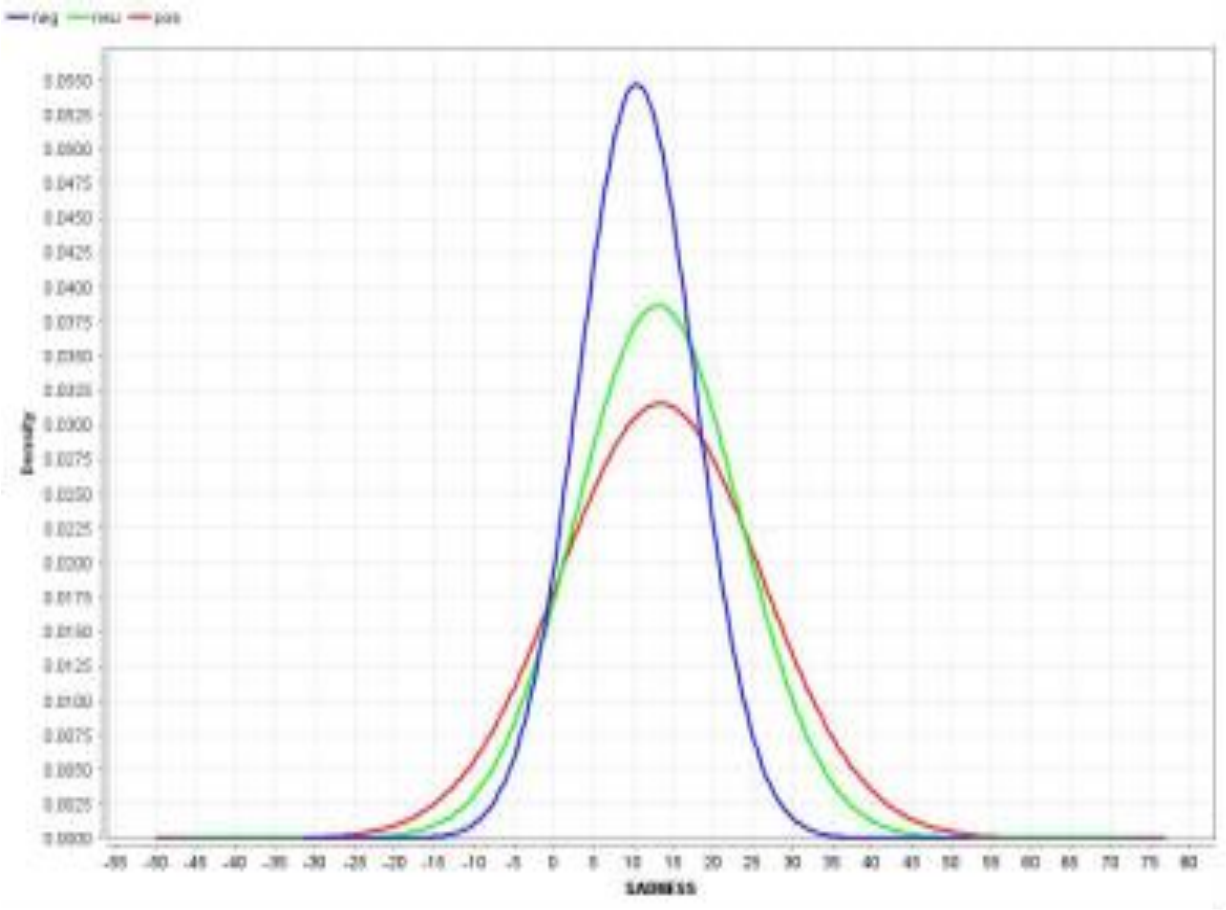

(c) SADNESS

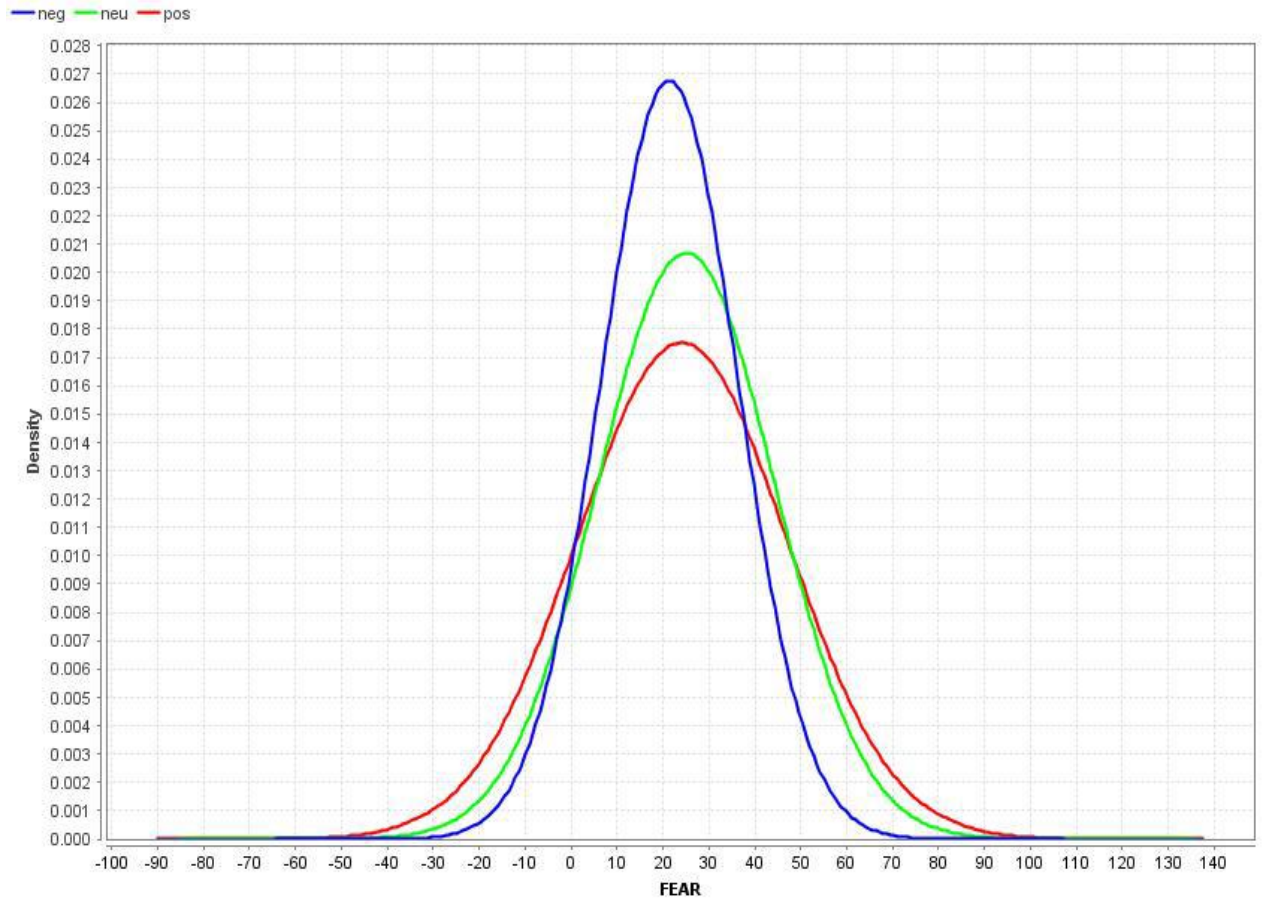

(d) FEAR

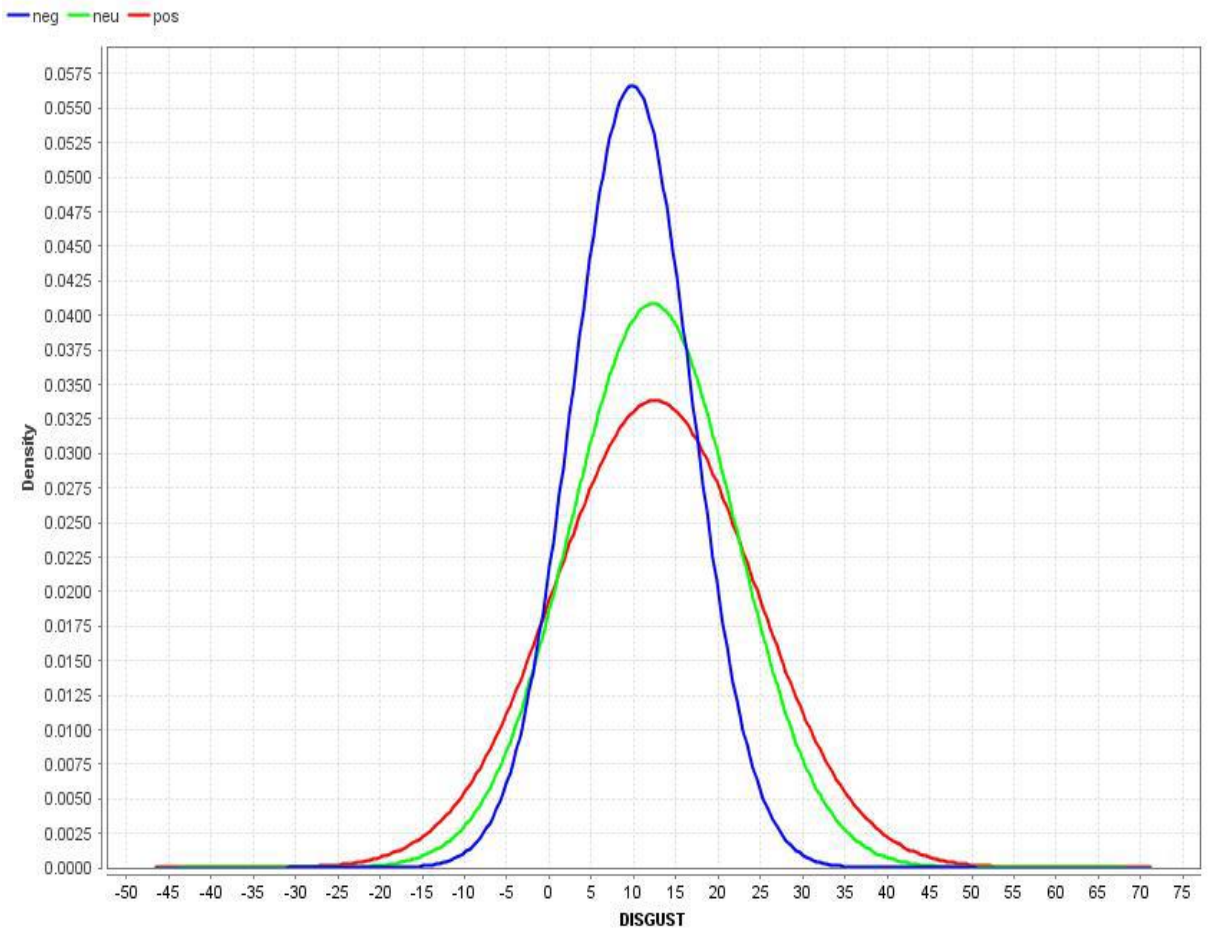

(e) DISGUST

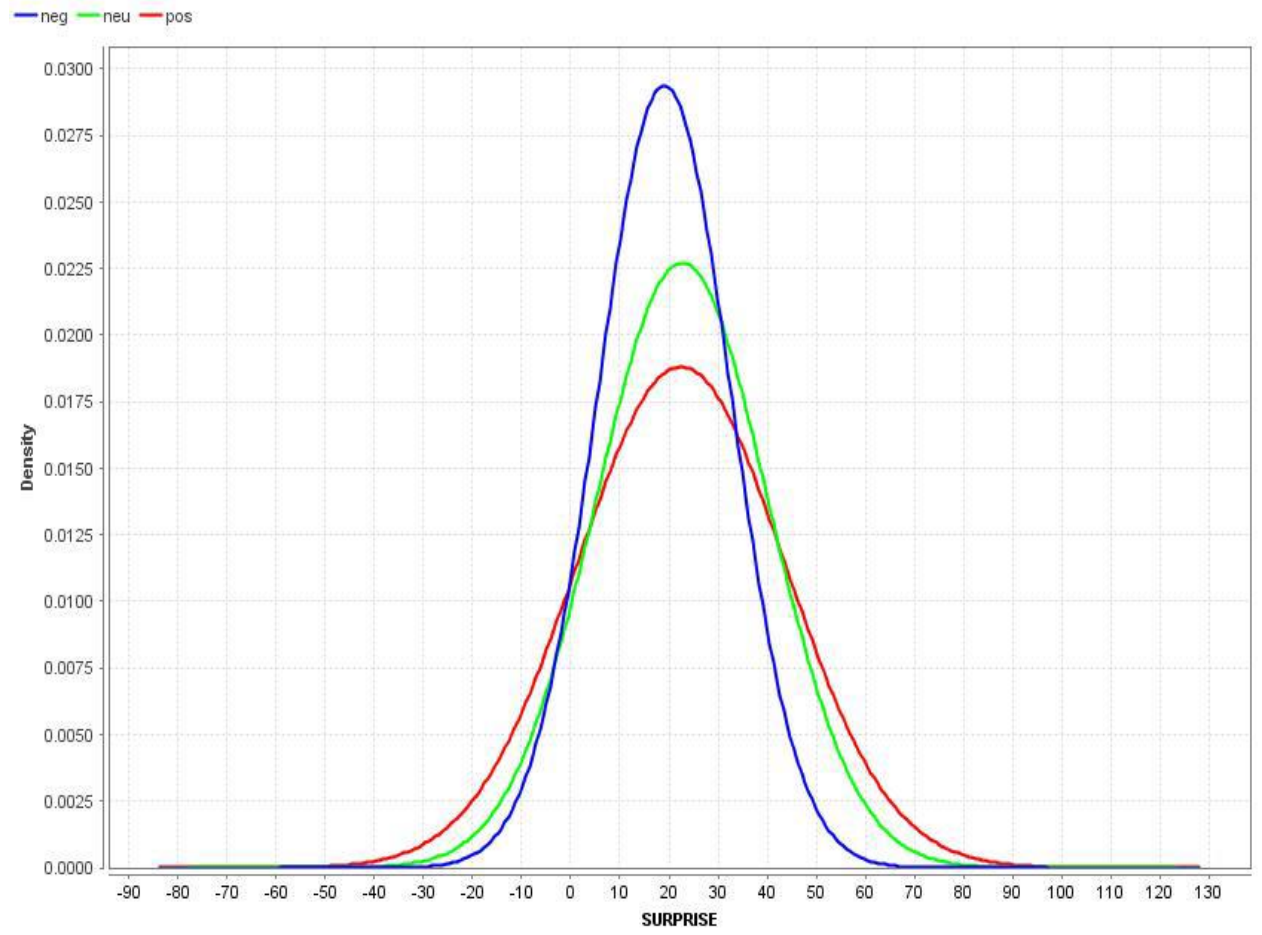

(f) SURPRISE

Figure 1: Distributions of emotions over **pos**, **neg** and **neu** document sentiments.

### 4.4. Word Senses

As the polarity of word changes with the senses and across domains [21, 22], it is important to know that in which sense each word is used in the sentence context [23]. For instance, consider the sentence S2 which is negative on "blow", whereas in sentence S3 it turns out to be positive. Having resolved all the word senses properly, the subsequent steps are going to become less error prone in determining the polarity of the sentence and document as well. S2: It came as a big blow to the project funding.

**S3**: The audio system will blow you away.

**S4**: A good beginner's tablet.

We did WSD using sense mapping service at idilia.com for entire document and analysed some of the samples and we found that the results are accurate in almost all cases. For instance, in the Sentence 4 the tablet is mapped to sense having the gloss "A graphics tablet is a computer peripheral device that allows one to hand-draw images directly into a computer, generally through an imaging program" which is good enough in identifying the concept related to laptops.

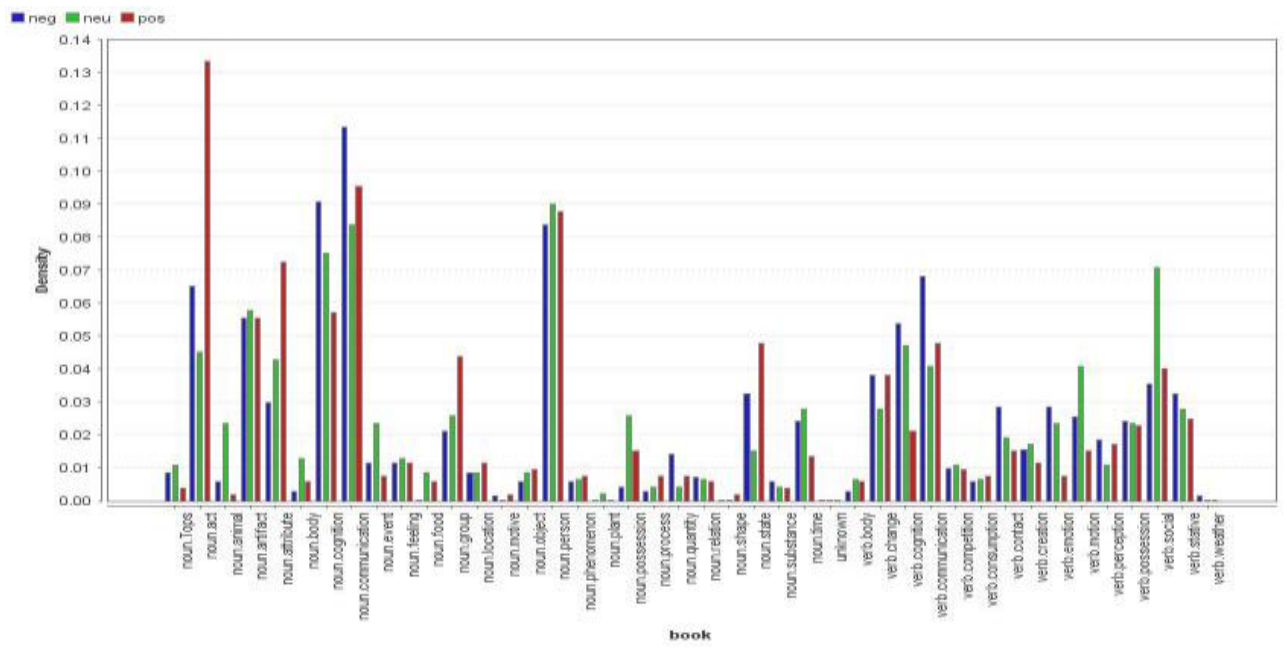

(a) BOOKS

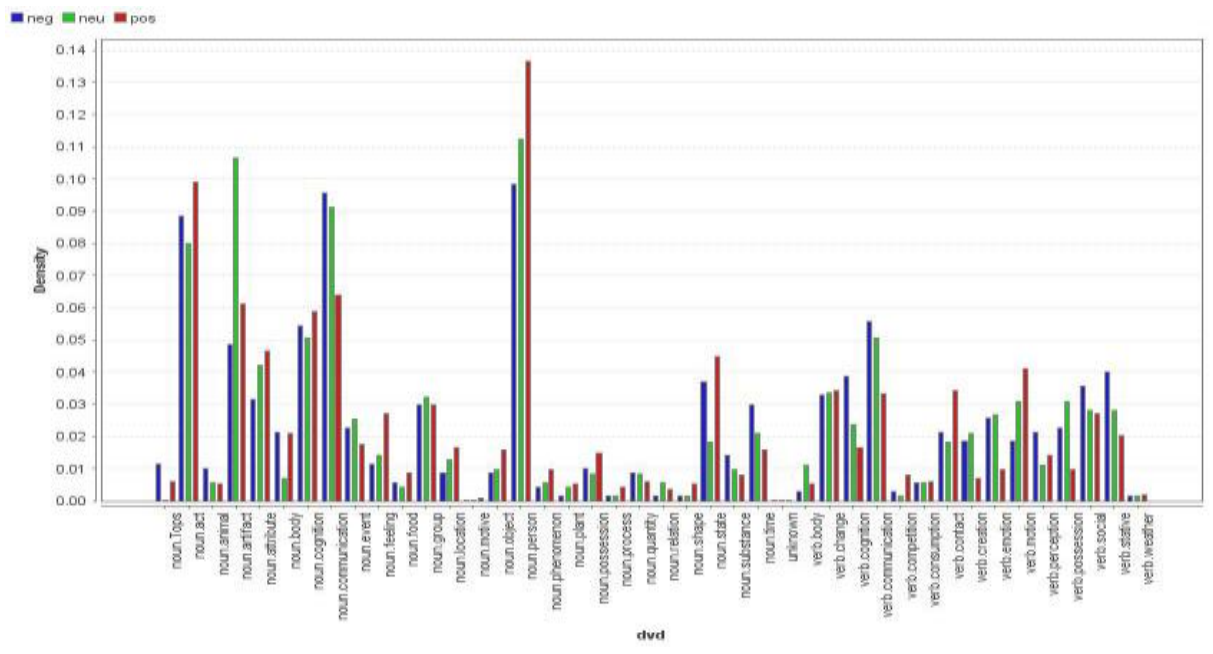

(b) DVD

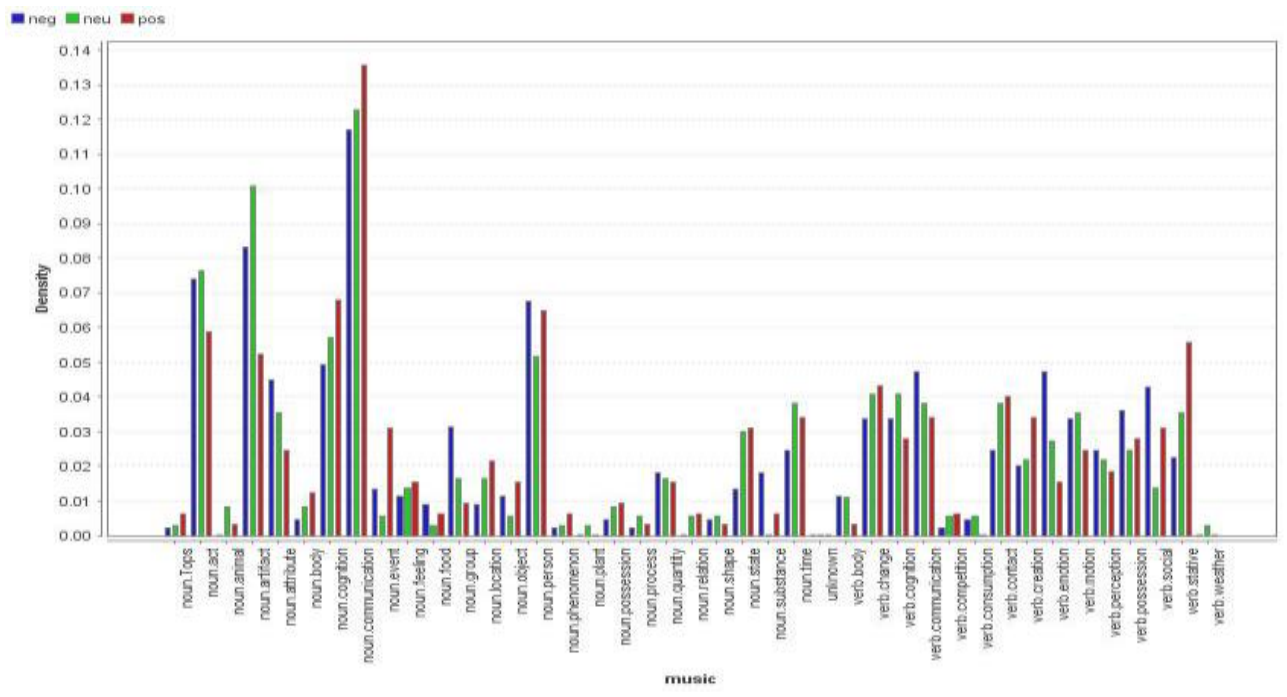

(c) MUSIC

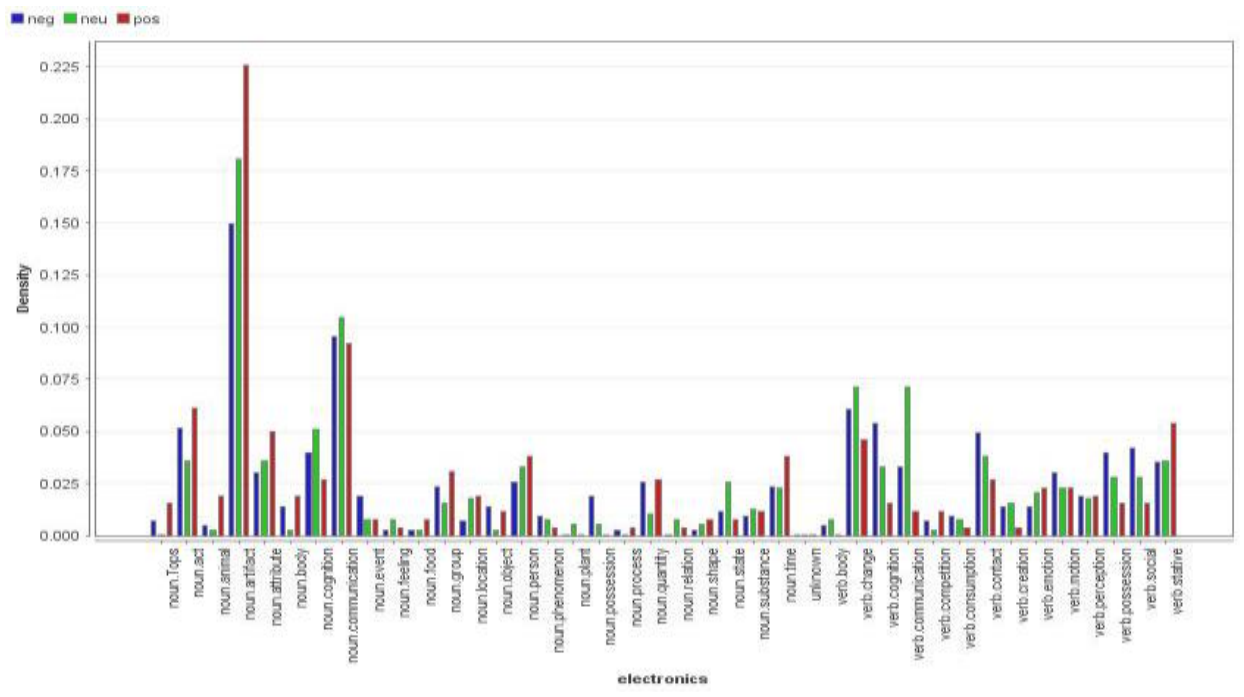

(d) ELECTRONICS

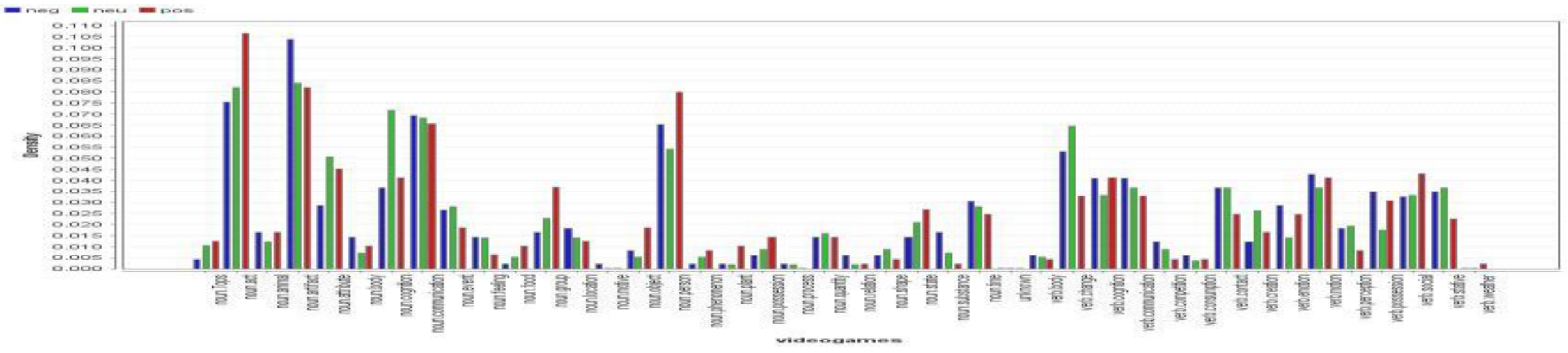

(e) VIDEOGAMES

Figure 2: Distributions of WordNet syntactic domains over **pos**, **neg** and **neu** document sentiments.

### 4.5. Lexicons

Most of the models in SA use a sentiment dictionary which is built automatically using a set of seed words [24] or a hand-built one. The automatically built dictionary such as SentiWordNet [25] consists of rich set of annotated sentiment words for translating words into polarity scores. For instance, consider the negative sentences S5, S6 and S7 which have negative phrases "**disproportionately**", "**put_to_sleep**" and "**one_dimensional**" respectively. Suppose using only the SentiWordNet gives neutral polarity for these words, which leads to misclassification into neutral class. In such cases, seeking for a second opinion or third to resolve the polarities of the words would be a better idea and will going to improve the accuracy. The bottom line is that seeking second opinion on the polarity of the words will ensure that the words which cannot be captured by a single dictionary could be captured by other dictionaries. When we crosschecked the words with other freely available dictionaries such as

a) **Lexicoder**[2] *(http://.lexicoder.com/download.html)*
b) **HarvardInquirer**[3] *(http://.wjh.harvard.edu/inquirer/spreadsheet_guide.htm)*
c) **SenticNet**[4] *(http://sentic.net/downloads)*
d) **Bing Liu Sentiment Lexicon**[5] *(https://www.cs.uic.edu/liub/FBS/)*
e) **MPQA Subjectivity Lexicon**[6] *(http://mpqa.cs.pitt.edu/#subj_lexicon)*
f) **SentiWords**[7] *(https://hltnlp.fbk.eu/technologies/sentiwords)*
g) **WordStat Sentiment Lexicon**[8] *(http://provalisresearch.com/products/)*

We found that majority of these dictionaries are resolving the words with correct polarities with which polarities of sentences could be determined correctly.

**S5**: The book is disproportionately focused on multilayer feedback networks.

**S6**: First of all, this book put me to sleep several times.

**S7**: This is a very one dimensional book.

### 4.6. Meta-Characters

A careful perusal of the input text data revealed that so many meta-characters such as **{:, ?, !, :D,** /} are intertwined with the sentences. The most frequently occurred meta- character ":" as used in the sentences S8 and S9 could be exploited in assessing the polarities of the sentiments expressed in the sentences. As per the syntax of the ":", the preceding expression of ":" will provide a clue about the following expression. In the sentences S10 and S11, the meta-character "/" is used to specify the overall rating which can be useful in evaluating the sentiments on the product. The 1/10 is negative the 5/10 is neutral and the 10/10 is positive ratings.

**S8:** The Good**:** it's a fun co-op game SP game (never tried PvP).

**S9:** **Pros:** Good features, stylish looks. **Cons:** This phone is very unreliable.

**S10:** Story**:** 1/10 here’s where things start to go bad.

**S11:** Sounds and Music**:** 7/10 Well, the sound is fantastic!

Table 3: Lexicographer file numbers and file names.

| FileNo. | Part-of-Speech.suffix | Description |
|---|---|---|
| 00 | adj.all | all adjective cluster |
| 01 | adj.pert | relational adjectives (pertainyms) |
| 02 | adv.all | all adverbs |
| 03 | noun.Tops | unique beginner for nouns |
| 04 | noun.act | nouns denoting acts or actions |
| 05 | noun.animal | nouns denoting animals |
| 06 | noun.artifact | nouns denoting man-made objects |
| 07 | noun.attribute | nouns denoting attributes of people and objects |
| 08 | noun.body | nouns denoting body parts |
| 09 | noun.cognition | nouns denoting cognitive processes and contents |
| 10 | noun.communication | nouns denoting communicative processes and contents |
| 11 | noun.event | nouns denoting natural events |
| 12 | noun.feeling | nouns denoting feelings and emotions |
| 13 | noun.food | nouns denoting foods and drinks |
| 14 | noun.group | nouns denoting groupings of people or objects |
| 15 | noun.location | nouns denoting spatial position |
| 16 | noun.motive | nouns denoting goals |
| 17 | noun.object | nouns denoting natural objects (not man-made) |
| 18 | noun.person | nouns denoting people |
| 19 | noun.phenomenon | nouns denoting natural phenomena |
| 20 | noun.plant | nouns denoting plants |
| 21 | noun.possession | nouns denoting possession and transfer of possession |
| 22 | noun.process | nouns denoting natural processes |
| 23 | noun.quantity | nouns denoting quantities and units of measure |
| 24 | noun.relation | nouns denoting relations between people or things or ideas |
| 25 | noun.shape | nouns denoting two and three dimensional shapes |
| 26 | noun.state | nouns denoting stable states of affairs |
| 27 | noun.substance | nouns denoting substances |
| 28 | noun.time | nouns denoting time and temporal relations |
| 29 | verb.body | verbs of grooming, dressing and bodily care |
| 30 | verb.change | verbs of size, temperature change, intensifying, etc. |
| 31 | verb.cognition | verbs of thinking, judging, analysing, doubting |
| 32 | verb.communication | verbs of telling, asking, ordering, singing |
| 33 | verb.competition | verbs of fighting, athletic activities |
| 34 | verb.consumption | verbs of eating and drinking |
| 35 | verb.contact | verbs of touching, hitting, tying, digging |
| 36 | verb.creation | verbs of sewing, baking, painting, performing |
| 37 | verb.emotion | verbs of feeling |
| 38 | verb.motion | verbs of walking, flying, swimming |
| 39 | verb.perception | verbs of seeing, hearing, feeling |
| 40 | verb.possession | verbs of buying, selling, owning |
| 41 | verb.social | verbs of political and social activities and events |
| 42 | verb.stative | verbs of being, having, spatial relations |
| 43 | verb.weather | verbs of raining, snowing, thawing, thundering |
| 44 | adj.ppl | participial adjectives |

## 5. CONCLUSION

SA is indeed a complex process which integrates several associated linguistic and statistical information and beyond the word features. In this paper, we have done some statistical and linguistic analysis on the opinionated text and based on the results and observations obtained in the analysis, the incorporation of the relevant features in feature vector would certainly enhance the system performance. There is a scope for further investigation into WordNet syntactic domains so as to utilize the syntactic domain information effectively but not limited to SA.

**Authors**

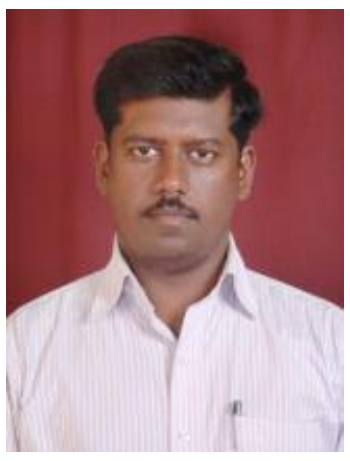

**K.PARAMESHA**
Assoc. Professor
Dept. of CSE, VVCE,
Mysore, India.

He has studied BE in Computer Science (Bangalore university) and M.Tech in Network Engineering (VTU, Belagavi). Presently working as a faculty in the Dept. Of Computer Science at VVCE, Mysuru and also actively involved in research on Sentiment Analysis. His primary focus is on Feature Engineering in Text Analytics. His areas of interest are Programming, NLP, Text Mining and Machine Learning.

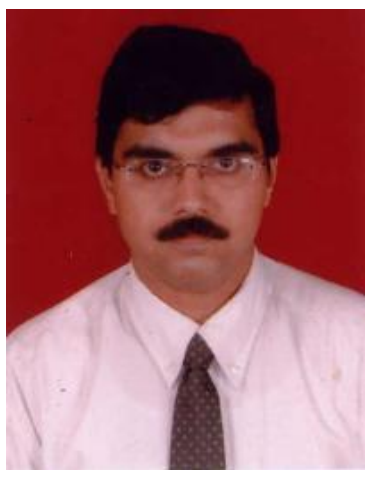

**Dr K.C.RAVISHANKAR**
Professor & HOD
Dept. of CSE,
Govt. Engineering College
Hassan, India

He has obtained BE in Computer Science and Engineering from Mysore University, M.Tech in Computer Science and Engineering from IIT, Delhi and doctorate from VTU, Belagavi in Computer Science and Engineering. He has 23 years of teaching experience. He served as Professor at Malnad College of Engineering Hassan. Presently working as Professor & HOD at Govt. Engineering College Hassan, Karnataka. His area of interest include Image Processing, Information security, Computer Network and Databases. He has presented a number of research papers at National and International conferences. Delivered invited talks and key note addresses at reputed institutions. He has served as chairperson and reviewer for conferences and currently guiding three PhD scholars.